\documentclass[letterpaper, 10 pt, conference]{ieeeconf}  
\pdfoutput=1
\usepackage{amsmath}
\usepackage{amssymb}
\usepackage{mathtools}
\usepackage{algorithm}
\usepackage{algpseudocode}
\usepackage[dvipsnames]{xcolor}
\usepackage{tabularx}
\usepackage{caption}
\usepackage{subcaption}
\newtheorem{remark}{Remark}
\usepackage{comment}

\IEEEoverridecommandlockouts                              
\title{\LARGE \bf
RL-based Variable Horizon Model Predictive Control of Multi-Robot Systems using Versatile On-Demand Collision Avoidance
}

\author{Shreyash Gupta$^1$, Abhinav Kumar$^1$, Niladri S. Tripathy$^2$ and Suril V. Shah$^1$
\thanks{$^{1}$Shreyash Gupta, Abhinav Kumar
 and Suril V. Shah are with the Department of Mechanical Engineering, Indian Institute of Technology Jodhpur,
        Jodhpur, India
      {\tt\small gupta.50@iitj.ac.in}; {\tt\small kumar.288@iitj.ac.in};  {\tt\small surilshah@iitj.ac.in}}%
\thanks{$^{2}$Niladri S. Tripathy is with the Department of Electrical  Engineering, Indian Institute of Technology Jodhpur,
        Jodhpur, India
        {\tt\small niladri@iitj.ac.in}}%
 }
       
\begin{document}
\maketitle
\thispagestyle{empty}
\pagestyle{empty}

\begin{abstract}
Multi-robot systems have become very popular in recent years because of their wide spectrum of applications, ranging from surveillance to cooperative payload transportation. Model Predictive Control (MPC) is a promising controller for multi-robot control because of its preview capability and ability to handle constraints easily. The performance of the MPC widely depends on many parameters, among which the prediction horizon is the major contributor. Increasing the prediction horizon beyond a limit drastically increases the computation cost. Tuning the value of the prediction horizon can be very time-consuming, and the tuning process must be repeated for every task. Moreover, instead of using a fixed horizon for an entire task, a better balance between performance and computation cost can be established if different prediction horizons can be employed for every robot at each time step. Further, for such variable prediction horizon MPC for multiple robots, on-demand collision avoidance is the key requirement. We propose Versatile On-demand Collision Avoidance (VODCA) strategy to comply with the variable horizon model predictive control. We also present a framework for learning the prediction horizon for the multi-robot system as a function of the states of the robots using the Soft Actor-Critic (SAC) RL algorithm. The results are illustrated and validated numerically for different multi-robot tasks.
\end{abstract}

\section{Introduction}
Multi-robot systems (MRS) are a collection of autonomous or semi-autonomous robots that can coordinate and work together to achieve a common goal. They are increasingly popular in various applications such as search and rescue, exploration, surveillance, and logistics \cite{Intro1}\cite{Intro2}. It can provide better efficiency, increased flexibility, and improved reliability compared to single robots in these applications \cite{Intro3}. However, controlling a group of robots presents a unique set of challenges, such as communication, task allocation, and coordination. To overcome these challenges, one must devise control strategies to guarantee that each robot can perform its tasks effectively and efficiently while maintaining overall system performance. Model Predictive Control (MPC) is considered one of the suitable controllers for multi-robot systems due to its ability to handle complex tasks, nonlinear dynamics and constraints. MPC is a control technique that uses a predictive model of the system and an optimization algorithm to generate control actions that minimize a cost function while satisfying system constraints \cite{Intro4}. In multi-robot systems, MPC can be used to optimize the coordination and motion of the robots in real-time while ensuring collision avoidance  and efficient task allocation \cite{Intro5}. MPC can also handle uncertainties \cite{Intro6} such as sensor noise and robot failures, making it a robust and reliable control strategy for multi-robot systems. In MPC, tuning of various parameters is crucial in determining the controller's performance. Selecting improper tuning parameters can lead to unstable or inefficient control behavior, e.g., if the prediction horizon is too short or the weighting factors are chosen poorly, the control system may not accurately track the reference trajectory or may consume excessive amounts of energy. The tuning process can be time-consuming and finding optimal values of tuning parameters require domain expertise and systematic experimentation to find the best set of parameters.

The prediction horizon in MPC is the length of time over which the future behavior of the system is predicted to generate optimal control actions. The prediction horizon is an important parameter that affects the performance of MPC. A longer prediction horizon can result in more accurate predictions of the system's behavior, leading to better control performance. However, it requires more computational resources, leading to slower control response time and potential instability. On the other hand, a shorter prediction horizon can result in faster control response time but it may lead to less accurate predictions of the system's behavior and suboptimal control performance \cite{maxN}. This observation served as the impetus for the development of the adaptive or variable horizon model predictive control. Researchers have devised several techniques to decide the optimal value of the prediction horizon. In \cite{lit1}, \cite{lit2}, the value of the horizon was included as an optimization variable in the MPC scheme. However, this may make the optimization problem non-quadratic and can lead to an increase in computation load. The quality of a predicted reference signal was used to decide how far off into the future one can effectively define the original MPC problem \cite{lit3}. An iterative deepening search where stability criteria are checked on each iteration to identify the lowest stabilizing horizon is one of the heuristics-based approaches proposed in \cite{lit4}. Most of these methods are dependent on heuristic approaches, which may not be optimum for a wide spectrum of multi-robot applications. Hence, learning-based approaches can be advantageous in estimating the prediction horizon for every time-step. Reinforcement Learning was used in \cite{RLMPC}, to learn the prediction horizon for a single robot avoiding static obstacles and was used in \cite{RLMPC2} to tune other meta-parameters of MPC. Further, extending such an approach to multi-robot systems may also require to deal with static and dynamic obstacles. The on-demand collision avoidance approach is shown to be advantageous in dealing with dynamic obstacles within MPC \cite{ODCA}. However, in the presence of variable prediction horizon, its preview capability may get compromised. 

To the best of our knowledge, RL based variable horizon MPC for multi-robot systems with improved on-demand collision avoidance to negotiate both static and dynamic obstacles is not reported in the literature. This serves as the main motivation behind this work. The contributions of this paper can be summarized as:
\begin{enumerate}
    \item We develop a versatile on-demand collision avoidance method (VODCA) for variable horizon MPC.
    \item We propose a framework to learn the prediction horizons of multiple robots in presence of static and dynamic obstacles using reinforcement learning.
    \item Numerical simulations are presented for different multi-robot configurations to confirm our propositions.
\end{enumerate}
\section{Problem Description and Proposed Solution}
\subsection{Problem Description:} 
Consider an environment with $N$ robots, where each robot can be defined using a single integrator model as
\begin{equation}
\label{single integrator model}
    \boldsymbol{\Dot{x}}_i = \boldsymbol{A}_c\boldsymbol{x}_i+\boldsymbol{B}_c\boldsymbol{u}_i,
\end{equation}
where state and input matrices $\boldsymbol{A}_c \in \mathbb{R}^{2 \times 2} = \boldsymbol{0}$ and $\boldsymbol{B}_c \in \mathbb{R}^{2 \times 2} = \boldsymbol{I}$, respectively. (\ref{single integrator model}) can be discretized to obtain a discrete-time linear model as
\begin{equation}
\label{discrete model}
    \boldsymbol{x}_i[k+1] = \boldsymbol{A}\boldsymbol{x}_i[k]+\boldsymbol{B}\boldsymbol{u}_i[k]
\end{equation}
where $\boldsymbol{A}\in \mathbb{R}^{2 \times 2} =\boldsymbol{I}$, $\boldsymbol{B} \in \mathbb{R}^{2 \times 2} = h\boldsymbol{I}$, $\boldsymbol{I}$ is an identity matrix, $h$ is the sampling time and $k$ is the discrete time-step.

Given the multiagent system consisting of $N$ robots with static and dynamic obstacles, the work addresses the following two problems :
\begin{itemize}
    \item \textbf{Problem 1:} Develop a method to ensure on-demand collision avoidance of $N$ robots with variable horizon Model Predictive Control (MPC).
    \item \textbf{Problem 2:} Learn the optimal sequence of the prediction horizon for all the $N$ robots to complete their task avoiding collision with static and dynamic obstacles.
  
\end{itemize}

\subsection{Proposed Solution:} 
In scenarios involving multiple robots, the ability to detect and prevent potential collisions is crucial. However, the current on-demand collision avoidance method \cite{ODCA} was insufficient to guarantee collision avoidance because the length of the prediction horizon ($N_i^k$) of $i^{th}$ robot at $k^{th}$ step varies at each time-step. To solve Problem 1, we propose VODCA (Versatile On-demand Collision Avoidance), a generalization of the on-demand collision avoidance technique, to guarantee the safe operation of multi-robot systems. The VODCA has the ability to ensure collision avoidance with variable horizon MPC, by exploiting the preview capability of MPC. Based on the values of prediction horizons of individual robots at individual time steps, the VODCA partitions the collision avoidance issue into three cases. These cases are then solved individually to generate collision avoidance constraints. The method is discussed in detail in Section \ref{VODCA section}.

To solve Problem 2, we use Soft Actor-Critic (SAC) RL algorithm to learn the optimal sequence of prediction horizon for each robot in a multi-robot system. We aim to learn a sequence, $\{N_i^0, N_i^1, N_i^2,\dots\}$ for each robot $i$, as a function of the states of the robots. 
We use a single Neural Network (NN) to learn the prediction horizon sequence of all the robots. The RL framework learns online with immediate rewards acquisition. The policy network takes in the states of all the robots and outputs a pair of mean and standard deviation for each robot. This Gaussian distribution, mean and standard deviation are passed through the hyperbolic tangent function to confine the value between -1 and 1; this value then is scaled and rounded off to obtain the prediction horizons of all the robots, $\{N_1^k,N_2^k,\dots,N_N^k\}$, at that time step $k$.

The next section discusses the variable horizon MPC formulation and the proposed collision avoidance method, VODCA.

\section{Variable prediction horizon MPC with a versatile on-demand collision}
In this section, we present variable prediction horizon MPC for multi-robot systems with a versatile on-demand collision avoidance strategy.
\subsection{The Prediction Model}
The prediction model for $i^{th}$ robot can be given by,
\begin{equation}\label{eq1}
    \hat{\boldsymbol{x}}_i[k_t + 1|k] = \boldsymbol{A}\hat{\boldsymbol{x}}_i[k_t|k] + \boldsymbol{B}\hat{\boldsymbol{u}}_i[k_t|k]
\end{equation}
where $k$ is current time step, $\hat{(\cdot)}[k_t|k]$ represents the predicted value of $(\cdot)[(k_t+k)h]$ with available information at $kh$, $k_t \in \{0,1,2,\dots,N_i^k-1\}$ and $h$ is the sampling time. $N_i^k$ represents the length of the prediction horizon of $i^{th}$ robot at $k^{th}$ time-step and it varies with change in $k$. Using (\ref{eq1}), for the  prediction horizon $N_i^k$, the prediction model can be written in the stacked form as
\begin{equation}
    \hat{\boldsymbol{X}}_i = \boldsymbol{F}\bar{\boldsymbol{x}}_i[0|k] + \boldsymbol{\Phi}\hat{\boldsymbol{U}}_i,
\end{equation}
where $\hat{\boldsymbol{X}}_i \in \mathbb{R}^{2N_i^k}$ and $\hat{\boldsymbol{U}}_i \in \mathbb{R}^{2N_i^k}$ are the stacked predicted states and inputs respectively, and $\bar{\boldsymbol{x}}_i[0|k]$ is the current measured state of the $i^{th}$ robot. The matrices $\boldsymbol{F} \in \mathbb{R}^{2N_i^k \times 2}$ and $\boldsymbol{\Phi} \in \mathbb{R}^{2N_i^k \times 2N_i^k}$ are given as
\begin{equation}
    \boldsymbol{F} = \begin{bmatrix}
         \boldsymbol{A}  \\
         \boldsymbol{A}^2 \\
         . \\
         . \\
         . \\
         \boldsymbol{A}^{N_i^k}
    \end{bmatrix}, \boldsymbol{\Phi} = \begin{bmatrix}
        \boldsymbol{B} & \boldsymbol{0} & \cdot & \boldsymbol{0} \\
        \boldsymbol{AB} & \boldsymbol{B} & \cdot & \boldsymbol{0} \\
        . & . & . & . \\
        . & . & . & . \\
        . & . & . & . \\
        \boldsymbol{A}^{N_i^k-1}\boldsymbol{B} & \boldsymbol{A}^{N_i^k-2}\boldsymbol{B} & \cdot & \boldsymbol{B}
    \end{bmatrix}.
\end{equation}
Since the prediction horizon $N_i^k$ varies, the dimensions of matrices $\boldsymbol{F}$ and $\boldsymbol{\Phi}$ will change with time.
\subsection{Cost Function}
We use a cost function comprising two terms: the first term minimizes the error between the current and desired states, and the second term minimizes the control input.  Let the desired state vector is denoted as  $\boldsymbol{X}_i^d \in \mathbb{R}^{2N_i^k}$. 
It is worth noting that for variable prediction horizon MPC the size of $\boldsymbol{X}_i^d$ will depend on the length of $N_i^k$. 
The cost function can then be represented as
\begin{equation}
J_i^{MPC}= (\boldsymbol{X}^d_i-\boldsymbol{\hat{X}}_i)^\top \boldsymbol{Q} (\boldsymbol{X}^d_i-\boldsymbol{\hat{X}}_i)
+ \boldsymbol{\hat{U}_i}^\top \boldsymbol{W} \boldsymbol{\hat{U}_i},
\end{equation}
where $\boldsymbol{Q} \in \mathbb{R}^{2N_i^k \times 2N_i^k} \geq 0$ and $\boldsymbol{W} \in \mathbb{R}^{2N_i^k \times 2N_i^k} > 0$ are the weight matrices. The optimization problem solves for the minimization of $J_i^{MPC}$ and yields optimal control inputs, ensuring the maneuver of robots from initial to desired states. In the next section, we discuss the proposed generalized collision avoidance method for multi-robot systems with variable prediction horizons.
\subsection{Versatile On-demand Collision Avoidance (VODCA)} \label{VODCA section}
On-demand collision avoidance schemes \cite{ODCA,CCTA} take advantage of the preview capability of MPC, look into the future predictions of respective states of the robots, and calculate the collision avoidance constraints only when it detects a collision in the future. However, the above schemes for the fixed horizon MPC when applied to variable prediction horizon MPC may suffer from a lack of preview capability when $N_i^{(k-1)}>N_i^k$. Therefore, we develop \textbf{V}ersatile \textbf{O}n-\textbf{D}emand \textbf{C}ollision \textbf{A}voidance (\textbf{VODCA}), the generalized version of On-demand collision avoidance \cite{ODCA} for variable prediction horizon MPC to ensure collision-free maneuver of multiple robots towards their desired state. 

The collision is detected by checking the error between the predicted states of $i^{th}$ robot, $\boldsymbol{\hat{X}}_i^{(k-1)} \in \mathbb{R}^{2N_i^{(k-1)}}$ and the predicted states of $j^{th}$ robot, $\boldsymbol{\hat{X}}_j^{(k-1)} \in \mathbb{R}^{2N_j^{(k-1)}}$ or the $l^{th}$ obstacle, $\boldsymbol{O}_{b,l}$. The possibility of $N_i^{(k-1)} \neq N_j^{(k-1)}$, makes this error calculation unfeasible. To overcome this problem, the VODCA first finds 
\begin{equation}
N_{min}^{(k-1)} = min(N_i^{(k-1)}, N_j^{(k-1)}).
\end{equation}
Next, we resize the length of the predicted state vector to $2N_{min}^{(k-1)}$. Now consider that robots and static obstacles are modeled as a disc with a radius equal to $r_{min}/2$. Then, the minimum distance required for collision avoidance and safe maneuvering of the robots in the environment is $r_{min}$.  The collision detection is performed based on the predicted states obtained at the previous time-step $(k-1)$. It can be said that there would be a collision in the future, if and only if, the distance ($d_i$) between the $i^{th}$ the robot and $j^{th}$ robot (which serves as a dynamic obstacle (DO)) or the $i^{th}$ robot and $l^{th}$ static obstacle (SO) satisfies the following:
\begin{equation}
\label{calculate_di}
d_{i}= 
\begin{cases}
	\left\|\boldsymbol{\hat{x}}_{i}[k_{c,i}|k-1]-\boldsymbol{\hat{x}}_{j}[k_{c,i}|k-1]\right\|< {r}_{min} & \text{If DO}\\
    \left\|\boldsymbol{\hat{x}}_{i}[k_{c,i}|k-1]-{\boldsymbol{o}}_{b,l}\right\|< {r}_{min} &\text{If SO}
\end{cases}
\end{equation}
 where, $k_{c,i}(\leq N_{min}^{(k-1)})$ is the time-step when the robot $i$ detects a collision with another robot or static obstacle. Therefore, in order to avoid collision, $i^{th}$ robot must attain a position at $(k_{c,i}-1+k)^{th}$ time-step such that,
\begin{equation}
\label{collision main equation}
\begin{cases}
	\left\|\boldsymbol{\hat{x}}_{i}[k_{c,i}-1|k]-\boldsymbol{\hat{x}}_{j}[k_{c,i}|k-1]\right\|\geq {r}_{min} & \text{If DO}\\
    \left\|\boldsymbol{\hat{x}}_{i}[k_{c,i}-1|k]-{\boldsymbol{o}}_{b,l}\right\|\geq {r}_{min} &\text{If SO}
\end{cases}
\end{equation}
is satisfied. Equation (\ref{collision main equation}) can be linearized using Multivariate Taylor-Series expansion about $\boldsymbol{\hat{x}}_{i}[k_{c,i}|k-1]$ and can be rearranged to obtain \cite{ODCA}\cite{CCTA}
\begin{equation}
\label{unstacked collision equation}
    {\boldsymbol{b}}_i^\top \boldsymbol{\hat{x}}_{i}[k_{c,i}-1|k] \geq c_i,
\end{equation}
where,
\begin{equation}
\label{calculate_bi}
{\boldsymbol{b}}_i= 
\begin{cases}
	(\boldsymbol{\hat{x}}_{i}[k_{c,i}|k-1]-\boldsymbol{\hat{x}}_{j}[k_{c,i}|k-1]) & \text{If DO}\\
    (\boldsymbol{\hat{x}}_{i}[k_{c,i}|k-1]-{\boldsymbol{o}}_{b,l}) &\text{If SO}
\end{cases}
\end{equation}
and $c_{i}=(r_{min}d_{i} - d_{i}^2)+\boldsymbol{b}_{i}^\top \boldsymbol{\hat{x}}_{i}[k_{c,i}|k-1]$. According to \cite{ODCA}, placing constraints on $\boldsymbol{\hat{x}}_{i}[k_{c,i}|k]$ rather than $\boldsymbol{\hat{x}}_{i}[k_{c,i}-1|k]$ (i.e. one time-step after the observed collision) improves collision avoidance even more. Stacked form of eq. (\ref{unstacked collision equation}) can be obtained by re-writing the eq. using $\boldsymbol{\hat{X}}_i^k \in \mathbb{R}^{2N_i^k}$ as,
\begin{equation}
\label{stacked collision equation}
    \boldsymbol{g}_i^\top \boldsymbol{\hat{X}}_i^{k} \geq c_i
\end{equation}
where $\boldsymbol{\hat{x}}_{i}[k_{c,i}|k]$ is the $k_{c,i}^{th}$ vector of $\boldsymbol{\hat{X}}_i^k$.

As mentioned earlier, in the case of the variable horizon MPC when $N_i^{(k-1)}>N_i^k$, it may happen that $k_{c,i}$ is greater than $N_i^k$. In such cases, (\ref{unstacked collision equation}) cannot be written in the stacked form given in (\ref{stacked collision equation}). This established the requirement of VODCA.
\begin{remark}
 The value of $k_{c,i}$ will always be less than or equal to $N_i^{k-1}$, as the collision is being predicted on the basis of information available at $(k-1)$ time-step. In fixed horizon MPC, $N_i^k$ is equal to $N_i^{k-1}$ and hence $k_{c,i}$ will also be less than or equal to $N_i^{k}$; but in variable horizon MPC, $N_i^k$ is  not necessarily equal to $N_i^{k-1}$, therefore, a clear relation between $k_{c,i}$ and $N_i^k$ cannot be established. 
\end{remark}
To deal with this, VODCA breaks down the variable $\boldsymbol{g}_i$ as follows
\begin{equation}\label{er13}
     \boldsymbol{g}_i^\top  =
    \begin{cases}
      \begin{bmatrix} \boldsymbol{b}_{i}^\top  & \boldsymbol{0}^{\top}_{2(N_i^k-1)\times  1}  \end{bmatrix}^\top & \text{ for $k_{c,i} = 1$}\\
      \begin{bmatrix} \boldsymbol{0}^{\top}_{2(k_{c,i}-1)\times  1} & \boldsymbol{b}_{i}^\top  & \boldsymbol{0}^{\top}_{2(N_i^k k_{c,i})\times  1} \end{bmatrix}^\top & \text{for $k_{c,i}<N_i^k$}\\
      \begin{bmatrix}	\boldsymbol{0}^{\top}_{2(N_i^k-1)\times  1} & \boldsymbol{b}_{i}^\top \end{bmatrix}^\top & \text{for $k_{c,i}=N_i^k$}
    \end{cases} 
\end{equation}
 Using (\ref{er13}), (\ref{stacked collision equation}) can be rearranged to obtain the following inequality constraints,
\begin{equation}
\label{collision avoidance constraints}
	\boldsymbol{A}_{i}^{coll}\hat{\boldsymbol{U}}_i \leq \boldsymbol{b}_{i}^{coll},
\end{equation}
where $\boldsymbol{A}_{i}^{coll}=-\boldsymbol{g}_{i}^\top\boldsymbol{\Phi} \text{ and } \boldsymbol{b}_{i}^{coll}=\boldsymbol{g}_{i}^\top\boldsymbol{F}\bar{\boldsymbol{x}}_i[0|k]-c_{i}$. The bounds on the control input, which ensures that it within the physical limits of the robot, can be stated as,
\begin{equation}
\label{control input bounds}
    \boldsymbol{lb} \leq \boldsymbol{\hat{U}} \leq \boldsymbol{ub},
\end{equation}
where $\boldsymbol{lb} \in \mathbb{R}^{2N_i^k}$ and $\boldsymbol{ub} \in \mathbb{R}^{2N_i^k}$ are the lower and upper bounds, respectively. Algorithm 1 describes implementing the proposed VODCA. 

The collision avoidance constraints (\ref{collision avoidance constraints}) and the control input bounds (\ref{control input bounds}) are augmented and written in the form of inequality constraints as,
\begin{equation}
    \boldsymbol{A}_i^{ineq} \boldsymbol{\hat{U}}_i \leq \boldsymbol{b}_i^{ineq},
\end{equation}
where $\boldsymbol{A}_i^{ineq}= {\begin{bmatrix} {\boldsymbol{A}^{coll}_i}^\top & -\boldsymbol{I}^\top & \boldsymbol{I}^\top
\end{bmatrix}}^\top \in \mathbb{R}^{(4N_i^k + 1)\times 2N_i^k}$ 
and $\boldsymbol{b}^{ineq}_i = {\begin{bmatrix}
    {\boldsymbol{b}^{coll}_i}^\top & -\boldsymbol{lb}^\top & \boldsymbol{ub}^\top
\end{bmatrix}}^\top\in \mathbb{R}^{(4N_i^k+1)\times1}$.

The quadratic optimization problem for a multi-robot system with collision avoidance can be stated as
\begin{equation}
    \begin{aligned}
       \min_{\boldsymbol{\hat{U}_i}} \quad & J^{MPC}_i \\
       \textrm{subject to} \quad & \boldsymbol{A}_i^{ineq} \boldsymbol{\hat{U}}_i \leq \boldsymbol{b}_i^{ineq}
    \end{aligned}
\end{equation}
The solution to this optimization problem generates an optimal control input sequence that encourages robots to converge to their respective desired state without getting collided. For variable horizon MPC, the selection of a prediction horizon is key challenge and will be addressed in the subsequent section. 

\begin{algorithm}
\caption{Versatile On-demand Collision Avoidance (VODCA).}
\begin{algorithmic}[1]
\State \textbf{Initialize:} $N_i^{k-1}$, $N_i^{k}$, $r_{min}$, $\textbf{F}$, $\boldsymbol{\Phi}$, $\hat{\textbf{X}}[k - 1]$, $\boldsymbol{O_b}$, $\bar{\textbf{x}}[k]$
\State \textbf{Output:} $\boldsymbol{A}_{i}^{coll}$ and $\boldsymbol{b}_{i}^{coll}$
\State $\boldsymbol{d}_i^{all} \gets$ \texttt{calDist}$(\boldsymbol{\hat{X}},\boldsymbol{O_b},N_i^{k-1},N_i^{k})$
\If{$\boldsymbol{d}_i^{all} < r_{min}$} \do \\
\State $[d_i,k_{c,i},\boldsymbol{b}_i] \gets$ \texttt{colsnPredict}$(\boldsymbol{\hat{X}},\boldsymbol{O_b})$
\State $c_i \gets$ \texttt{calculateCi}$(r_{min},d_i,\boldsymbol{b}_i,\boldsymbol{\hat{X}}[k-1],k_{c,i})$
\If{$k_{c,i} = 1$}
    \State $\boldsymbol{g}_i^\top = \begin{bmatrix}
	\boldsymbol{b}_{i}^\top  & \boldsymbol{0}^{\top}_{2(N_i^k-1)\times  1}
    \end{bmatrix}^\top$
\ElsIf{$k_{c,i}<N_i^k$}
    \State $\boldsymbol{g}_i^\top = \begin{bmatrix}
	\boldsymbol{0}^{\top}_{2(k_{c,i}-1)\times  1} & \boldsymbol{b}_{i}^\top  & \boldsymbol{0}^{\top}_{2(N_i^k-k_{c,i})\times  1}
    \end{bmatrix}^\top$
\ElsIf{$k_{c,i}=N_i^k$}
    \State $\boldsymbol{g}_i^\top = \begin{bmatrix}
	\boldsymbol{0}^{\top}_{2(N_i^k-1)\times  1} & \boldsymbol{b}_{i}^\top \end{bmatrix}^\top$
\EndIf
\EndIf
\State $[\boldsymbol{A}_{i}^{coll},\boldsymbol{b}_{i}^{coll}] \gets$ \texttt{calColsnConst}$(\boldsymbol{g}_i,\boldsymbol{\Phi},c_i,\textbf{F},\bar{\textbf{x}}_i[0|k])$
\end{algorithmic}
\end{algorithm}

\section{LEARNING OF PREDICTION HORIZON}
In this work, we propose using RL to learn the prediction horizon for each robot. We employ Soft Actor-Critic (SAC) \cite{SAC}, an  entropy-maximization actor-critic RL algorithm with a parameterized stochastic policy, to determine the value of the prediction horizon at each time-step of MPC. In RL, the environment is defined as a Markov Decision Process (MDP), which is represented using a tuple $(\mathcal{S},\mathcal{A},\mathcal{P},\mathcal{R},\gamma)$. For the given environment, $\mathcal{S}$ is the set of all possible states, $\mathcal{A}$ is the set of all possible actions, $\mathcal{P}$ is the state transition function, $\mathcal{R}$ is the reward function, and $\gamma \in [0,1)$ is the discount factor which decides how future rewards are valued. The aim of RL is to come up with a policy $\pi_\theta$ such that it maximizes the expected sum of rewards acquired over the states visited by the policy in an episode, i.e., to optimize:
\begin{equation}
    J^{RL}(\theta) = \underset{\theta}{\max} \mathbb{E} \Biggr[\sum_{k=0}^{K} \gamma^k R(s,\pi_\theta(s))\Biggr], \forall s_0 \in \mathcal{S}_0.
\end{equation}
where $K$ is the maximum number of time steps and $\theta$ represents the design parameters of the policy network. The trajectory distribution produced by the policy and the state transition function, as well as the starting state distribution $\mathcal{S}_0$, are taken into account when calculating the expectation.

In SAC, an additional entropy term is augmented with the RL objective function defined as,
\begin{equation}
    \alpha\mathcal{H}(\pi_\theta (\cdot|s)) = \alpha\mathbb{E}_{a \sim \pi_\theta(a|s)} [-\text{log}(\pi_\theta(a|s))].
\end{equation}
This entropy term favors stochastic policies and encourages exploration of the environment. The intensity of exploration can be regulated by a temperature parameter $\alpha$ which determines the relative importance of the entropy term against the reward. For higher values of $\alpha$, the policy will be explorative in nature, and for lower values of $\alpha$, it will be exploitative in nature. 

We build on the result reported in \cite{RLMPC} to extend it to learn the prediction horizon of multiple robots in the presence of static and dynamic obstacles instead of a single robot avoiding static obstacles. Hence, the environment consists of total $N$ robots with dynamics (\ref{discrete model}), completing their tasks by avoiding obstacles using MPC with VODCA. The RL state vector $s\in\mathbb{R}^{2N}$ consists of measured positions $\boldsymbol{\bar{x}}_i[0|k]$, of all the robots. The policy function used is a stochastic policy that inputs $s$ and outputs a mean ($\mu_\theta$) and log-standard deviation ($\sigma_\theta$) of a Gaussian distribution. The policy network outputs a pair of mean and log-standard deviation for each robot. A normal distribution is then instantiated with the mean and standard deviation, and a sample is drawn from it using the reparameterization trick. Policy action is then obtained by transforming this sample using a hyperbolic tangent function to ensure it falls within the range [-1,1], implemented as:
\begin{equation}
    \pi_{\theta,i}(s) = tanh(\mu_{\theta,i} + \sigma_{\theta,i} \odot \mathcal{N}(0,1)),
\end{equation}
where $\pi_{\theta,i}(s)$ is the policy output corresponding to $i^{th}$ robot. The Gaussian policy allows for smooth and differential mapping, which is important for efficient gradient-based optimization during training. The prediction horizon is a natural number. It was observed in the literature \cite{maxN} that in robotic applications, a value of prediction horizon greater than 50 is rarely used; therefore, we choose an upper bound on the prediction horizon, $N_i^{max}$ equal to 49. The value of the prediction horizon for each robot can be estimated from $\pi_{\theta,i}(s)$ by first scaling the output from the $tanh$'s limits of -1 and 1, to 1 and $N_i^{max}$, and then rounding up the output to the next natural number:
\begin{equation}
    N_i^k = \text{ceiling}(\text{scale}(\pi_{\theta,i}(s),[-1,1],[1,N_i^{max}])),
\end{equation}
$N_i^k$ represents the prediction horizon of $i^{th}$ robot generated at $k^{th}$ time-step. It is reported in \cite{RLMPC} that the gradients of the RL problem are not affected by these transformations as they are applied in the environment, while the gradients are calculated based on the unscaled and unceilinged outputs from $\pi_{\theta, i}(s)$.

The Reward function consists of five terms. The propagation reward $R_{p}$, ensures the robots have maximum possible movement towards their goal positions on each step. The computation reward $R_{h}$, enables the network to generate the minimum prediction horizons. The collision reward $R_{c}$, penalizes collision among the robots and with the obstacles. The termination reward $R_{term}$, rewards every robot on reaching the vicinity of the goal within an error threshold $\bar{e}$. The variance reward $R_{v}$, is the last term that enforces the network to minimize variation among the prediction horizons of the robots and thus provides the equal ability to all the robots to predict the collisions. The proposed reward function is given below:
\begin{equation}
    R = \sum_{i=1}^{N} \Biggr[ R_{p,i} - R_{h,i} - R_{c,i} + R_{term,i} \Biggr] - R_{v}
\end{equation}

The propagation reward $R_{p,i}$ is designed in such a way that it provides a greater reward to a robot when it moves toward the goal and penalizes when it moves away, i.e., 
\begin{equation}
\label{calculate_bi}
R_{p,i}= 
\begin{cases}
	25 + 250\Delta e_i  & \text{If } \Delta e_i \geq 0\\
    250\Delta e_i  &\text{If } \Delta e_i < 0,
\end{cases}
\end{equation}
where $\Delta e_i = e_i^{(k-1)} - e_i^k$, the change in error between two consecutive time-steps. $e_i^{k} = \left\|\boldsymbol{x}_{i}^d-\boldsymbol{\bar{x}}_{i}[k]\right\|$ is the error between the desired and current position of the $i^{th}$ robot at $k^{th}$ time-step. A robot can be said to be moving towards its desired position if $e_i^{(k-1)} \geq e_i^k $. Next, $R_{h,i}$ can be defined as
\begin{equation}
    R_{h,i} = \lambda_h N_i^k,
\end{equation}
where $\lambda_h$ is a constant. $R_{c,i}$ is defined as
\begin{equation}
\label{calculate_bi}
R_{p,i}= 
\begin{cases}
	100 + (K - k) & \text{If collides}\\
    0 &\text{otherwise}.
\end{cases}
\end{equation}
Here, we are considering that a robot has collided when it comes in the vicinity of $0.5r_{min}$ with any other robot or obstacle. Further, $R_{v}$ is given as
\begin{equation}
    R_{v} = Var(N_i^k).
\end{equation}
Finally, $R_{term,i}$ is defined as
\begin{equation}
\label{calculate_bi}
R_{term,i}= 
\begin{cases}
	  0 & \text{If }  e_i^k \geq \bar{e}\\
    150 &\text{If }  e_i^k < \bar{e}.
\end{cases}
\end{equation}
\subsection{Implementation and Network Architecture}
The framework learns the prediction horizon online with immediate rewards at every time step. The critic network architecture consists of two fully connected layers with ReLU activation functions. The first layer takes in the concatenated state-action pair as input, and the output of the final layer is a single scalar value representing the soft Q-value. The soft Q-value formulation allows for a more robust training process by adding stochasticity to the target values. There are 256 nodes in each layer of the critic network. The actor architecture consists of two fully connected layers with ReLU activation functions. The first layer takes in the state as input, and the output of the final layer is split into two branches: one branch outputs the mean of the policy distribution, and the other outputs the logarithm of the standard deviation. The actor-network also has 256 nodes in each layer. We use two soft Q-value networks, and two target Q-value networks and take their minimum to estimate the soft Q-value better.
\section{RESULTS AND DISCUSSIONS}
\begin{figure}[h]
        \centering 
        \includegraphics[scale=0.45]{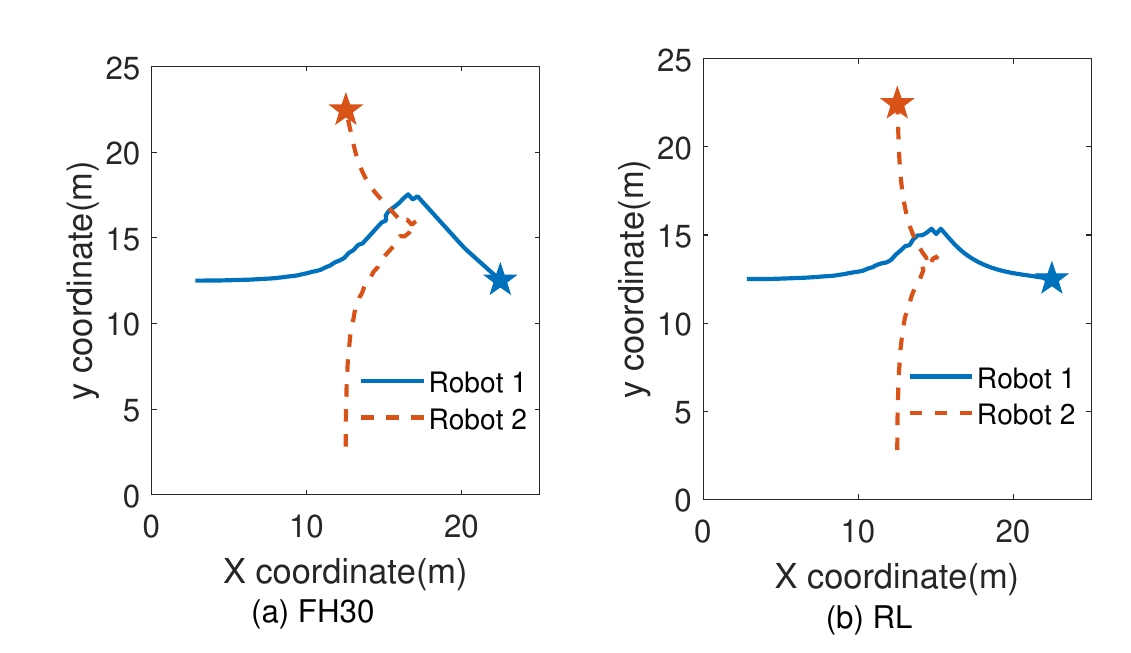}
        \caption{Set point control of two robots. It must be noted that the robots take a longer path in FH compared to RL.}
        \label{motion2R}
\end{figure}

\begin{figure}
    \centering
    \includegraphics[scale=0.45]{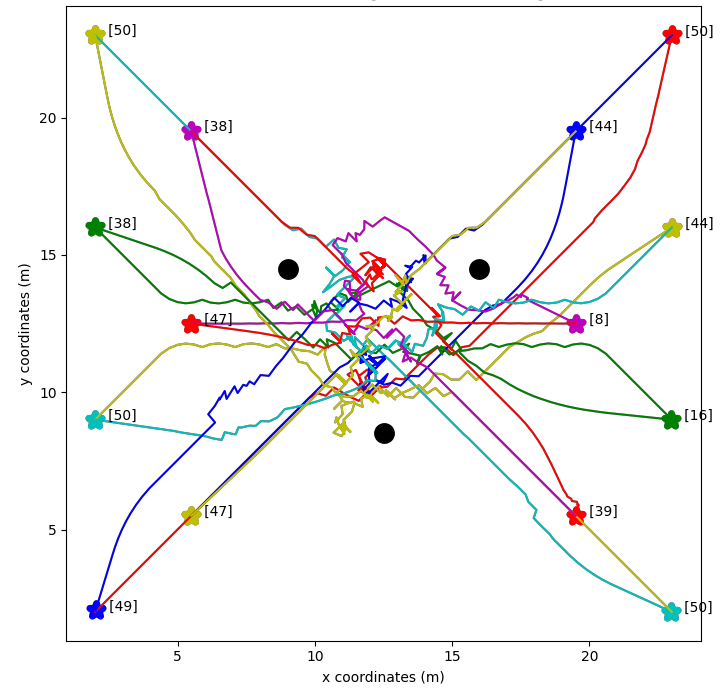}
    \caption{Motion plot of 14 robots exchanging their position in dynamic environment using RL based variable horizon MPC. The number in square brackets is the prediction horizon of the respective robot at that time-instant.}
    \label{14R motion plot}
\end{figure}

\begin{figure}[h]
    \begin{subfigure}[b]{0.4\columnwidth}
        \centering
        \includegraphics[scale=0.25]{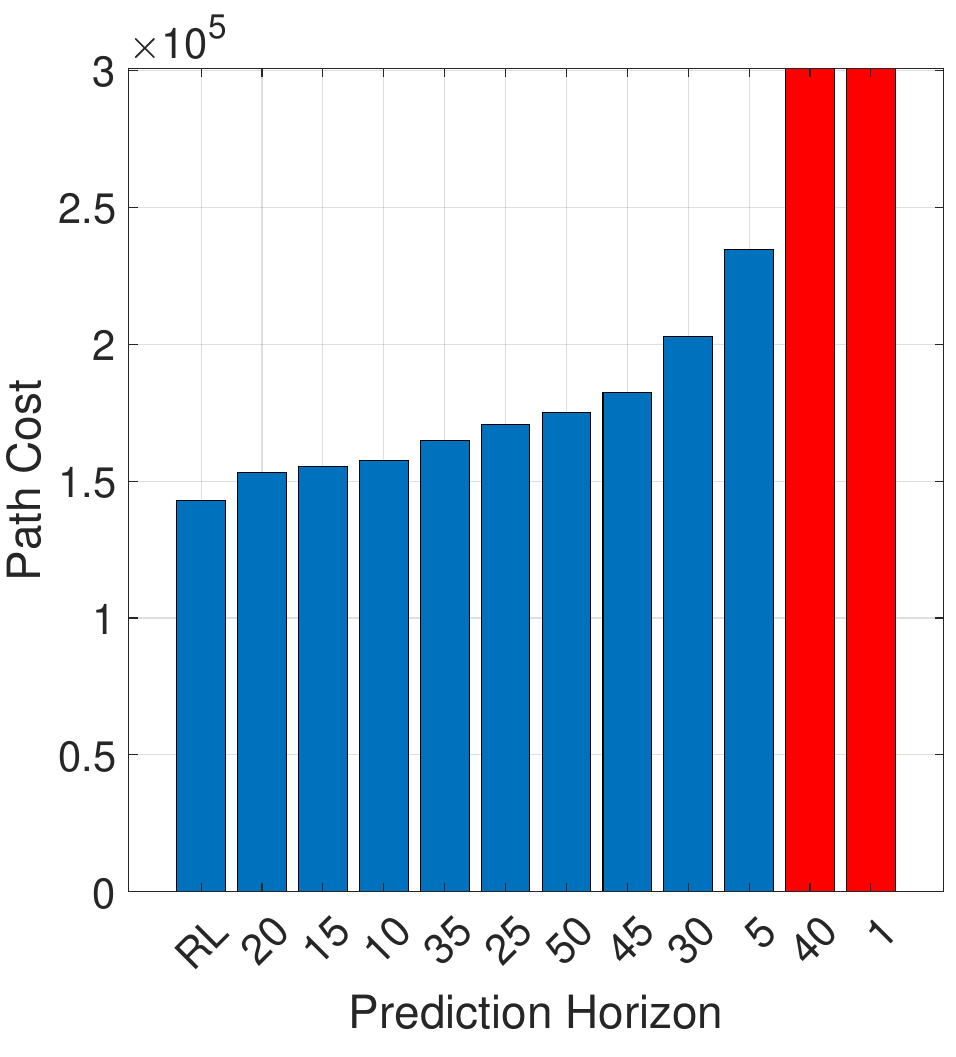}
        \caption{Path cost of different fixed horizon and RL.}
        \label{14R_SSSC}
    \end{subfigure}
    \hspace{0.7cm}
    \begin{subfigure}[b]{0.4\columnwidth}
        \centering
        \includegraphics[width=122pt,height=122pt]{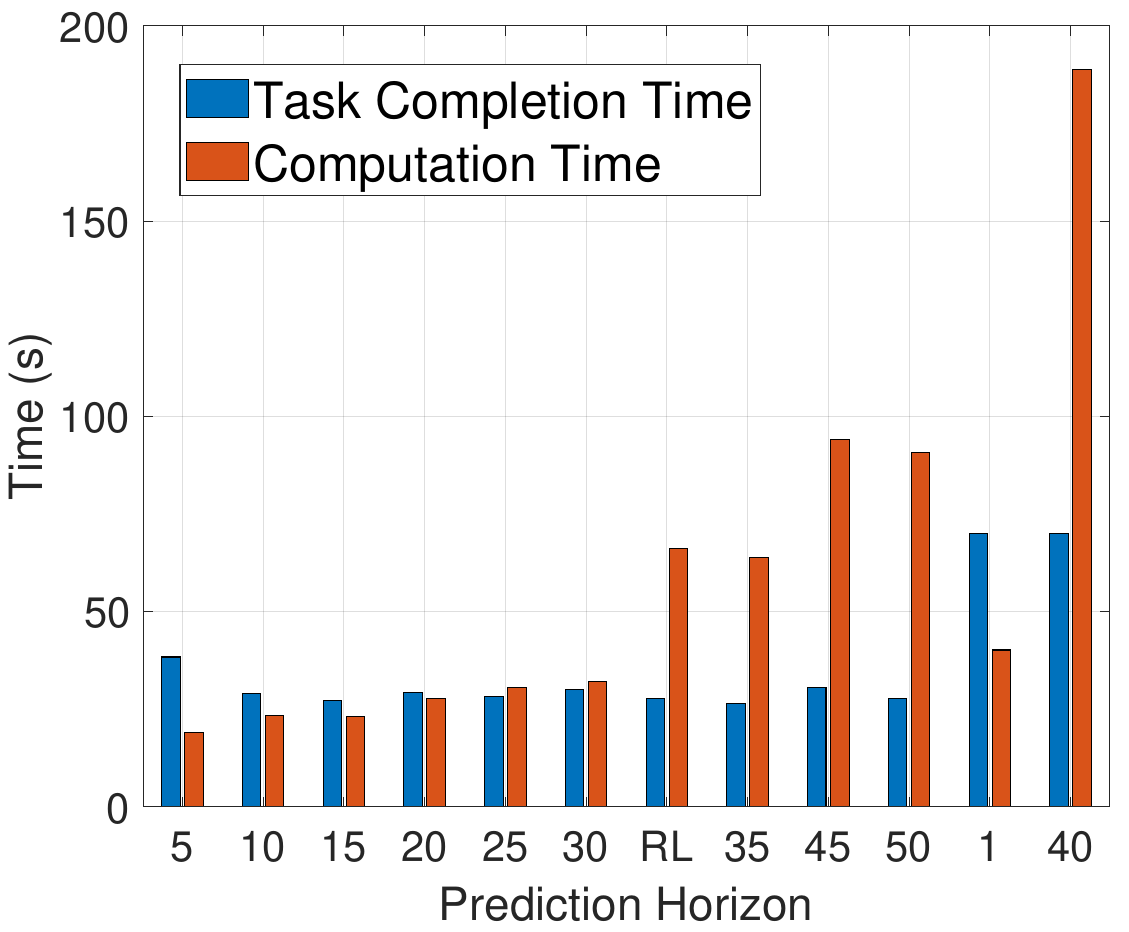}
        \caption{Computation time and task completion time.}
        \label{14R_Time_Bar_Plot}
    \end{subfigure}
        \caption{Comparison of performance cost and computation cost for different fixed prediction horizons and RL-based variable horizon. It can be noted here that RL-based variable horizon MPC is the best performing with an improvement of 7.07\% over the best fixed horizon. Here, the red colored bar depicts the fixed horizons which were not able to complete the task.}
        \label{2R_RL_Motion}
\end{figure}

The proposed framework was implemented on various multi-robot tasks like set-point control, position exchange, and segregation of multiple robots into the desired number of groups. All the simulations were run on a Linux-based Ubuntu 20.01 having intel core $i7$-$8^{th}$ Gen with $8GB$ RAM. The value of various parameters used in numerical studies are given in Table \ref{tab:1}, where $\boldsymbol{I}$ is the identity matrix of appropriate dimensions.
\begin{table}[h]
\caption{Input parameters}
\label{tab:1}
\begin{tabularx}{0.45\textwidth} { 
  | >{\centering\arraybackslash}X 
  | >{\centering\arraybackslash}X 
  | >{\centering\arraybackslash}X | }
 \hline
 $h = 0.2 sec$ & $\boldsymbol{Q} = 5\boldsymbol{I}$ & $\boldsymbol{W} = 10\boldsymbol{I}$ \\
 \hline
 $r_{min}=1.5m$  & $ub = 1.5m/s$  & $lb = -1.5m/s$  \\
\hline
 $K = 350$  & $\lambda_h = 0.001$  & $\bar{e} = 0.1$  \\
\hline
\end{tabularx}
\end{table}
For simplicity, we will use the abbreviation RL for the proposed RL based variable horizon MPC and FH for fixed horizon MPC. First, we present set point control of two robots to show the efficacy of the proposed framework. The initial positions of Robot-1 and -2 are (12.5, 2.5) and (2.5, 12.5), respectively, and the final positions are (12.5, 22.5) and (22.5, 12.5), respectively, all in meters. Fig.\ref{motion2R} represent the motion plots generated using RL and FH, respectively. 
\begin{figure}[t]
    \begin{subfigure}[b]{0.4\columnwidth}
        \centering
        \includegraphics[scale=0.21]{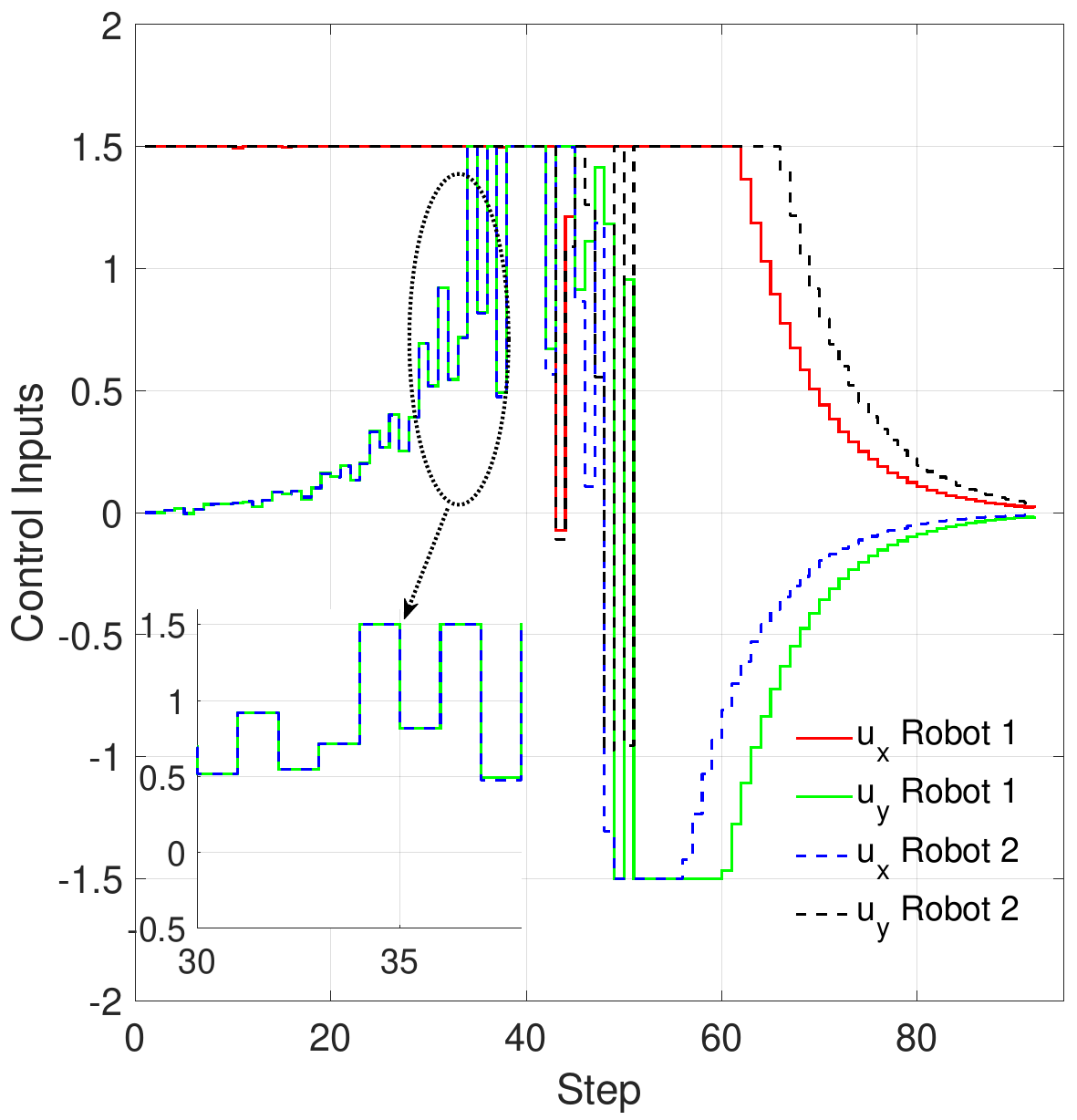}
        \caption{FH30}
        \label{Control inputs:FH30}
    \end{subfigure}
    \hspace{0.75cm}
    \begin{subfigure}[b]{0.4\columnwidth}
        \centering
        \includegraphics[scale=0.21]{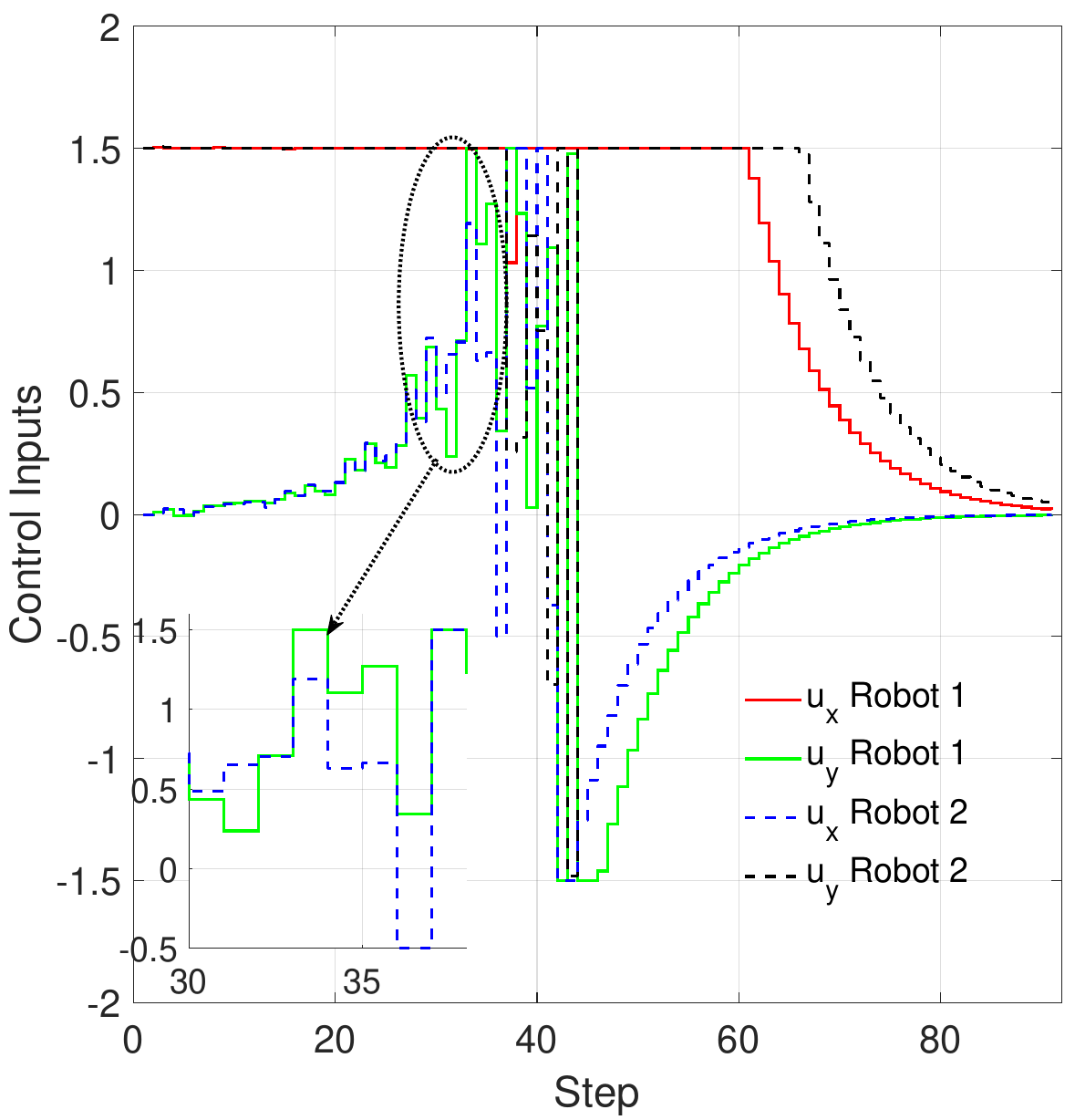}
        \caption{RL}
        \label{Control inputs:RL}
    \end{subfigure}
        \caption{Control inputs of both the robots for set point control. $u_{y,1}$ and $u_{x,2}$ are identical in the highlighted portion, in case of FH30. This results in more collision constraint activation and a longer path in this configuration.}
        \label{2R_RL_Motion}
\end{figure}
The RL takes a shorter path in comparison to the best-performing FH. This behaviour can be better understood by looking at the control inputs of both the robots in the case of RL and FH, shown in Fig. \ref{Control inputs:RL} and Fig. \ref{Control inputs:FH30}, respectively. In Fig. \ref{Control inputs:FH30}, it can be observed that the values of $u_{y,1}$ and $u_{x,2}$ (shown as solid green and dashed blue lines, respectively) are identical till time-step 40, which drives both the robots to their midpoint (12.5, 12.5) and they end up being into continuous collision avoidance tussle, which enforces them to take a longer path. In contrast, in RL (Fig. \ref{Control inputs:RL}), the values of $u_{y,1}$ and $u_{x_2}$ are not identical. This happens because of the variable prediction horizons at each time step. We want the robots to maneuver in a way that activates the collision avoidance constraints as minimally as possible, which will result in less computation as well as a smoother path. Due to the variable horizon, RL generated different magnitudes of control inputs for both the robots and activated the collision avoidance constraints close to a minimum number of times. 
\begin{figure}
    \begin{subfigure}[b]{0.4\columnwidth}
    \centering
        \includegraphics[scale=0.25]{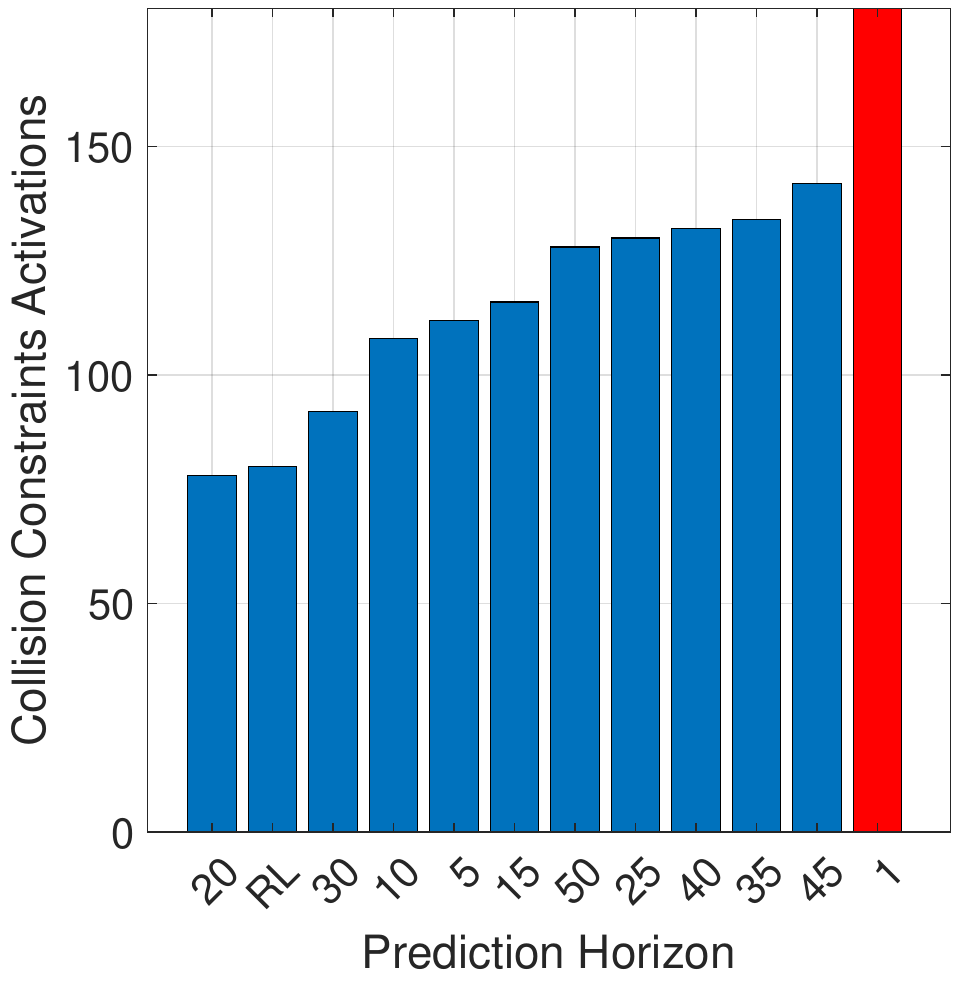}
        \caption{Number of collision constraints activation.}
        \label{Number of collision constraints activations}
    \end{subfigure}
    \hspace{0.75cm}
    \begin{subfigure}[b]{0.4\columnwidth}
        \centering
        \includegraphics[scale=0.25]{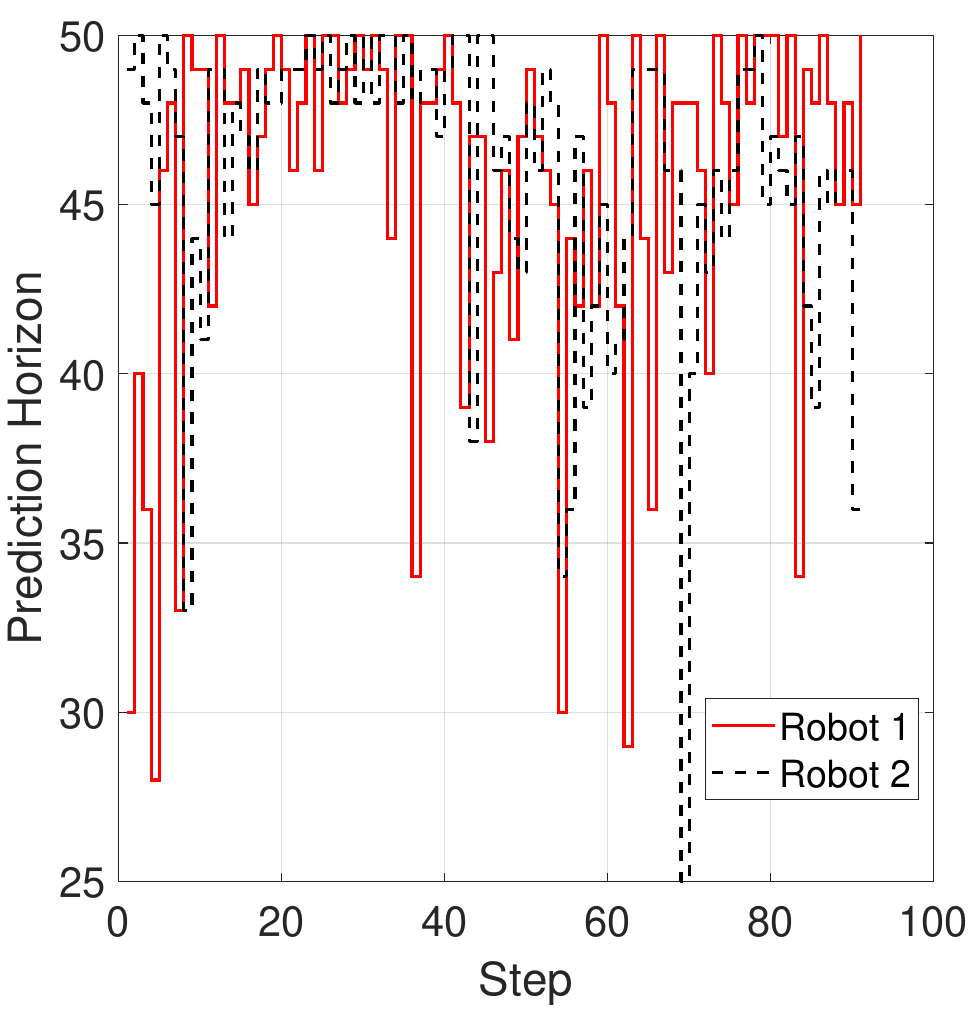}
        \caption{Prediction horizon wrt time-step.}
        \label{Prediction horizon wrt time-step}
    \end{subfigure}
        \caption{Number of collision constraint activation and prediction horizon propagation for set-point control task. Here, the red colored bar depicts the fixed horizon which was not able to complete the task.} 
        \label{2R_RL_Motion}
\end{figure}
The collision activations count for the different fixed horizons and RL is shown in Fig. \ref{Number of collision constraints activations}. Fig. \ref{Prediction horizon wrt time-step}, shows the propagation of prediction horizon over time-steps, for both the robots.
Next, a position exchange task was performed using 14 robots in the presence of three static obstacles. In Fig. \ref{14R motion plot}, seven robots started from the left side, and the other seven started from the right side and were required to exchange their positions. We have chosen path cost as the performance metric. Fig. \ref{14R_SSSC} compares the path cost of RL and various FHs. RL was best performing with an improvement of 7.07\% over the best performing fixed horizon, FH20. It must be noted that the path cost does not have a linear relation with the value of the prediction horizon. This clearly emphasizes the need for learning-based variable prediction horizon MPC, as it would be difficult to find out an optimal fixed horizon. CPU computation time and task completion time (= no. of iterations $\times h$) for various FHs and RL is also shown in Fig. \ref{14R_Time_Bar_Plot}. The task completion time of RL was comparable to best-performing FHs, but the computation time was higher in RL, which can be reduced using a better computing device. This case proves that the framework is scalable and can be used to perform different multi-robot tasks with many robots with equally good or better performance than FH. This shows the efficacy of the proposed framework. 

\begin{figure}[h]
    \centering
        \includegraphics[scale=0.3]{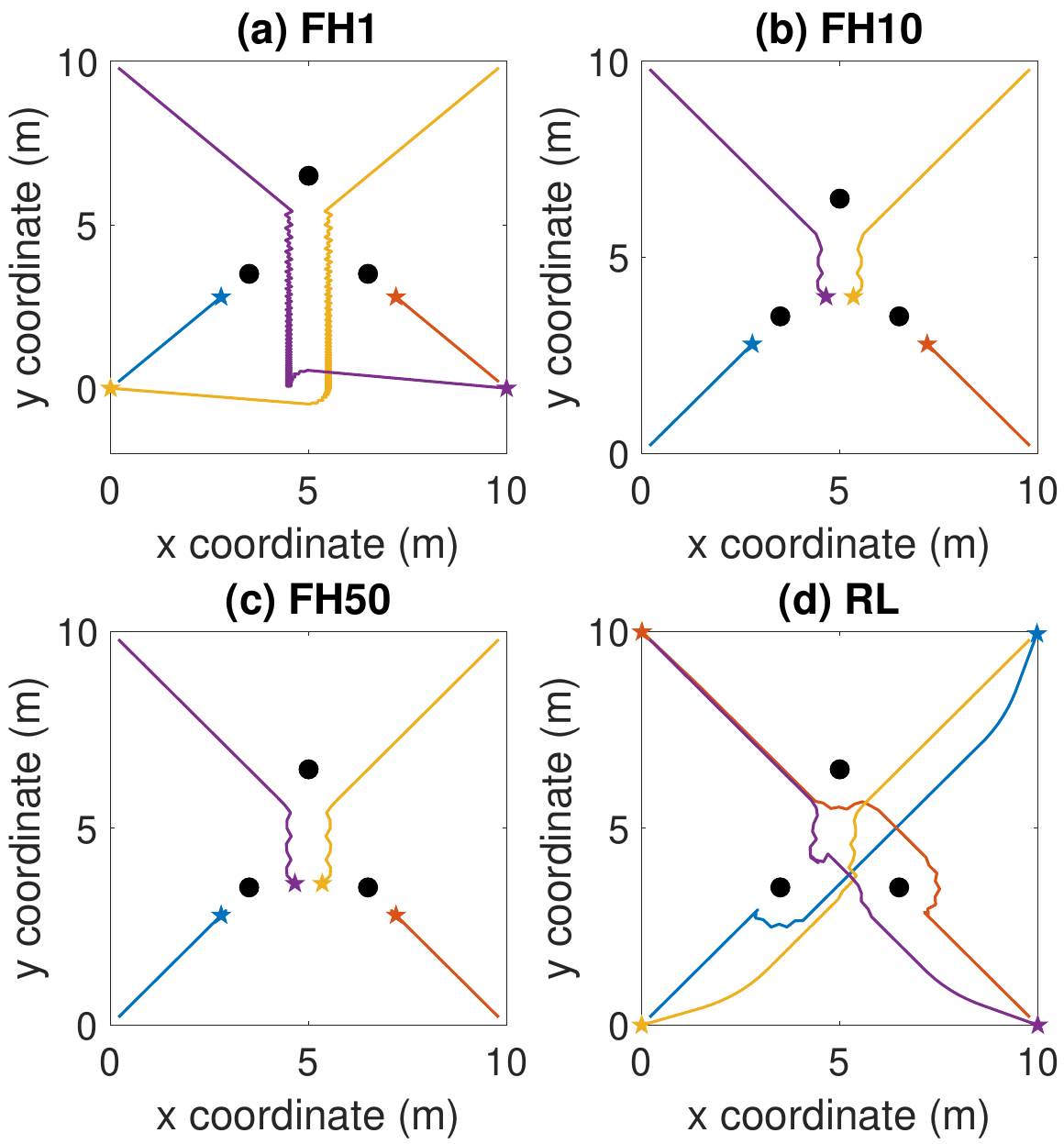}
        \caption{Motion plot of set-point control of four robots. Only RL based variable horizon MPC was able to complete the manoeuvre while avoiding the obstacles.}
        \label{only RL FH1}
\end{figure}

It is worth reporting that there were few configurations in which only RL could complete the task successfully. One such case is shown in Fig. \ref{only RL FH1}, when FH was used, robots got into local minima around a static obstacle or collided with other robots. But RL was able to find the path because of the different prediction horizons of every robot at each time step. It can be seen that the proposed RL-based variable prediction MPC for multi-robot systems was able to complete all the above-mentioned tasks with equally good or better performance than fixed horizon MPC. Moreover, it has some clear advantages over fixed horizon MPC like dealing with local minima and minimum collision constraints activation. Successful completion of all the discussed cases proves that the proposed variable horizon MPC with VODCA is scaleable to many robots.
\section{CONCLUSIONS AND FUTURE WORKS}
 In this work, we proposed Versatile on-demand collision avoidance (VODCA) for variable prediction horizon MPC. VODCA mitigates the drawback of on-demand collision avoidance in the form of limited preview capability when the length of prediction horizon, $N_i^{(k-1)} > N_i^k$. Further, we propose learning variable prediction horizon for multi-robot systems in presence of dynamic and static obstacles using the Soft Actor-Critic Algorithm. We introduced a novel reward function which was able to drive robots towards their desired positions, enforcing minimum prediction horizon and drastic change in its value. 

 In order to show the efficacy of the proposed framework, three different experiments are presented. First, a set point control of two robots was carried out where RL-based variable horizon MPC generated a smaller path with a minimum number of collision constraint activation. Next, a position exchange of fourteen robots was carried out in presence of static and dynamic obstacles. Here, for a given robot, other robots serve as dynamic obstacles. The simulation results showed that the robots were successfully able to exchange their positions and the proposed framework performs better than any fixed horizon MPC. Finally, an example  of four robot systems in a dynamic environment was presented where fixed horizon MPC fails but the proposed RL-based variable horizon MPC was able to find the solution. This shows the superiority of the proposed framework. Future work will focus on real hardware implementation and extending this framework for varied multi-robot tasks. 

\section{ACKNOWLEDGMENTS}

The authors gratefully acknowledge the contribution of Sudhir Pratap Yadav for insightful discussion on RL and Saurabh Chaudhary for helping in generating results.

\addtolength{\textheight}{-3cm}   


\begin{thebibliography}{99}

\bibitem{Intro1}
Dias MB, Zlot R, Kalra N, Stentz A. "Market-based multirobot coordination: A survey and analysis." Proceedings of the IEEE (2006):1257-70.

\bibitem{Intro2}
Rizk Y, Awad M, Tunstel EW. "Cooperative heterogeneous multi-robot systems: A survey." ACM Computing Surveys (2019):1-31.

\bibitem{Intro3}
Gautam A, Mohan S. "A review of research in multi-robot systems." IEEE 7th International Conference on Industrial and Information Systems (2012):1-5.

\bibitem{Intro4}
Camacho EF, Alba CB. "Model predictive control." Springer Science and Business Media (2013).

\bibitem{Intro5}
Zhu EL, Stürz YR, Rosolia U, Borrelli F. "Trajectory optimization for nonlinear multi-agent systems using decentralized learning model predictive control." IEEE Conference on Decision and Control (2020):6198-6203.

\bibitem{Intro6}
Gonzalez R, Fiacchini M, Guzmán JL, Alamo T. "Robust tube-based MPC for constrained mobile robots under slip conditions." IEEE Conference on Decision and Control held jointly with Chinese Control Conference (2009):5985-5990.

\bibitem{lit1}
Scokaert PO, Mayne DQ. "Min-max feedback model predictive control for constrained linear systems." IEEE Transactions on Automatic control (1998):1136-42.

\bibitem{lit2}
Richards A, How JP. "Robust variable horizon model predictive control for vehicle maneuvering." International Journal of Robust and Nonlinear Control: IFAC‐Affiliated Journal (2006):333-51.

\bibitem{lit3}
Droge G, Egerstedt M. "Adaptive time horizon optimization in model predictive control." American Control Conference (2011):1843-1848

\bibitem{lit4}
Krener AJ. "Adaptive horizon model predictive control." IFAC-PapersOnLine (2018):31-6.

\bibitem{RLMPC}
Bøhn E, Gros S, Moe S, Johansen TA. "Reinforcement learning of the prediction horizon in model predictive control." IFAC-PapersOnLine (2021):314-20.

\bibitem{RLMPC2}
Bøhn E, Gros S, Moe S, Johansen TA. "Optimization of the Model Predictive Control Meta-Parameters Through Reinforcement Learning." arXiv preprint arXiv:2111.04146 (2021).


\bibitem{ODCA}
Luis, Carlos E., and Angela P. Schoellig. "Trajectory generation for multiagent point-to-point transitions via distributed model predictive control." IEEE Robotics and Automation Letters (2019): 375-382.

\bibitem{maxN}
Schwenzer M, Ay M, Bergs T, Abel D. "Review on model predictive control: An engineering perspective." The International Journal of Advanced Manufacturing Technology (2021):1327-49.


\bibitem{CCTA}
Gupta S, Chaudhary S, Maurya D, Joshi SK, Tripathy NS, and Shah SV. "Segregation of Multiple Robots Using Model Predictive Control With Asynchronous Path Smoothing." IEEE Conference on Control Technology and Applications (2022): 1378-1383

\bibitem{SAC}
Haarnoja T, Zhou A, Abbeel P, Levine S. "Soft actor-critic: Off-policy maximum entropy deep reinforcement learning with a stochastic actor." International Conference on Machine Learning (2018): 1861-1870
\end{thebibliography}
\end{document}